\documentclass{article}

\usepackage{mystyle,times}
\usepackage{hyperref}
\usepackage{url}

\usepackage{epsfig}
\usepackage{graphicx}
\usepackage{amsmath}
\usepackage{amssymb}
\usepackage{enumitem}

\usepackage{microtype}
\usepackage{tikz}
    \definecolor{mycolor1}{HTML}{009900}
    \definecolor{mycolor2}{HTML}{990000}
    \definecolor{mycolor3}{HTML}{0000EE}
    \definecolor{green}{HTML}{009900}
    \definecolor{red}{HTML}{990000}
    \usetikzlibrary{decorations.pathmorphing}
	\tikzstyle{cut-edge}=[dotted]
    \tikzstyle{vertex}=[circle, draw, fill=white, inner sep=0pt, minimum width=1ex]
    \tikzset{every picture/.append style={baseline,scale=1.1}}
\usepackage[rightcaption]{sidecap}
    
\usepackage[hyperref, standard]{ntheorem}

\hypersetup{colorlinks, linkcolor=mycolor3, citecolor=mycolor3, urlcolor=mycolor3}

\renewcommand{\textcolor}[1]{}

\def\clap#1{\hbox to 0pt{\hss#1\hss}}

\def\mathclap{\mathpalette\mathclapinternal}

\def\mathclapinternal#1#2{%
\clap{$\mathsurround=0pt#1{#2}$}}

\title{Convexity Shape Constraints for Image Segmentation}

\newcommand*\samethanks[1][\value{footnote}]{\footnotemark[#1]}

\author{
Loic A Royer\textsuperscript{1}\thanks{Shared first authors}, David L Richmond\textsuperscript{1}\samethanks, Carsten Rother\textsuperscript{2}, \\
\textbf{Bjoern Andres\textsuperscript{3}, and Dagmar Kainmueller\textsuperscript{1}} 
\vspace{3pt} \\
\vspace{2pt}
\textsuperscript{1}MPI-CBG, \textsuperscript{2}TU Dresden, \textsuperscript{3}MPI for Informatics\\
\texttt{kainmueller@mpi-cbg.de} 
}

\nipsfinalcopy 
\begin{document}

\maketitle

\begin{abstract}
Segmenting an image into multiple components is a central task in computer vision.
In many practical scenarios, prior knowledge about plausible components is available.
Incorporating such prior knowledge into models and algorithms for image segmentation is highly desirable, yet can be non-trivial.
In this work, we introduce a new approach that allows, for the first time, to constrain some or all components of a segmentation to have convex shapes.
Specifically, we extend the Minimum Cost Multicut Problem by a class of constraints that enforce convexity.
To solve instances of this APX-hard integer linear program to optimality, 
we separate the proposed constraints in the branch-and-cut loop of a state-of-the-art ILP solver.
Results on natural and biological images demonstrate the effectiveness of the approach 
as well as its advantage over the state-of-the-art heuristic.
\end{abstract}


\section{Introduction}

Image segmentation is a challenging task for which often times the use of suitable prior knowledge about the shape of the sought objects plays an important role. 
One interesting shape prior is convexity~\cite{Strekalovskiy2011,Veksler,RotherCVPR2010,BoykovECCV2014}. 
In natural images, it often occurs that there are multiple convex structures of the same or different classes present in one image, such as wheels or various fruit, as well as composite structures constructed from convex parts, such as bricks and floor tiles. 
Biology similarly gives numerous examples of multiple convex structures. For example, many cell types are convex, such as bacteria, yeast, and more complicated cells densely packed into tissue. Numerous subcellular structures, including nuclei and various types of vesicles, are also convex. 
Despite the clear relevance of this situation to the task of image segmentation, respective priors have, to the best of our knowledge, not yet been addressed in the literature. 

Existing methods for segmenting multiple convex structures are specifically designed for certain shapes like ellipsoids or rods, as e.g.\  \cite{delgado2012snakes}.  
Such methods do not enforce generic convexity, but instead employ priors of specific shapes that happen to be convex. 
Furthermore, such methods commonly segment multiple structures sequentially,  
and neither the reconstruction of individual structures nor the resulting segmentation of multiple structures is globally optimal w.r.t.\ the underlying objective. 

Beyond specific shape priors, there has been recent interest in generic convexity priors for binary image segmentation. 
\emph{Star convexity} priors were introduced in~\cite{Veksler}, where convexity is defined with respect to all rays emanating from a central, user-defined seed point. This approach was generalized to the case of Geodesic Star Convexity by~\cite{RotherCVPR2010}, which defines convexity with regards to Geodesic paths. 
Truly convex objects were first handled by~\cite{Strekalovskiy2011}. 
However, this approach is limited to a single foreground class, which must be explicitly modeled. 
The task of segmenting convex foreground objects without the requirement of user input or explicit modeling is studied in~\cite{BoykovECCV2014}. They propose a graphical model with triple-cliques that encode convexity constraints as 1-0-1 label sequences along straight lines in the image. This formulation captures the global nature of convexity. However, it implies that only one connected component of the foreground class can be present. Furthermore, the complexity of the problem requires the use of approximate solvers 
which may lead to local minima. 
None of the above methods is able to segment many generic convex objects of multiple foreground classes, fully automatically, without the need for user-defined seed points.

\begin{figure*}
\begin{center}
   \includegraphics[width=0.92\textwidth]
                   {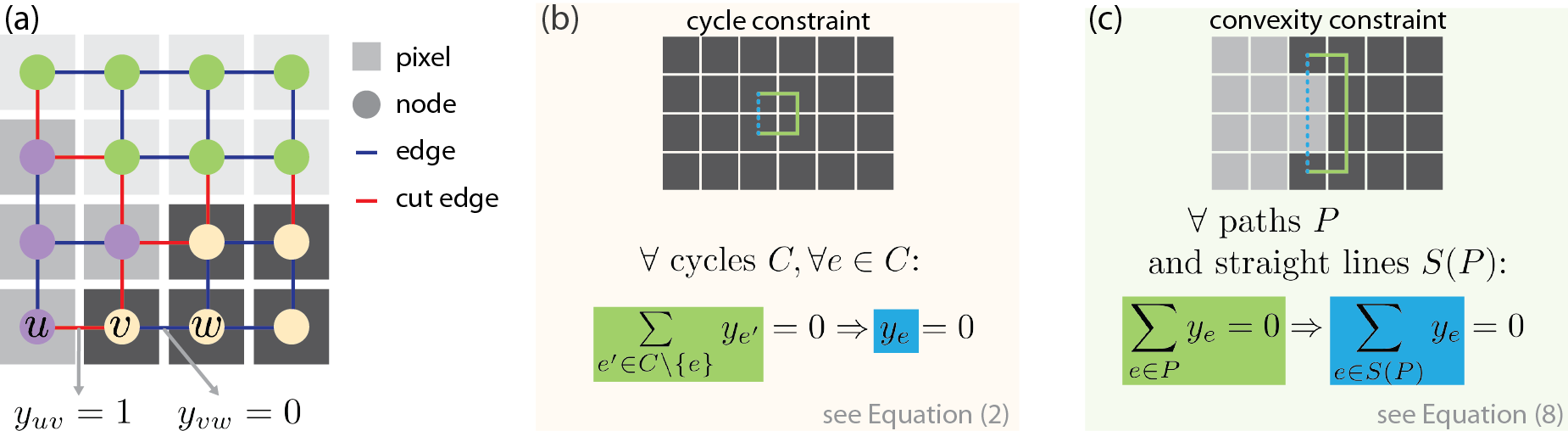} 
\end{center}
\vspace{-3pt}
   \caption{Method overview. (a)~The Multicut Problem \ref{def:multicut} expresses image decomposition as a binary edge labeling problem in a pixel grid graph. 
	Components are inferred from the edge labels via connectivity analysis. (b)~Cycle constraints enforce that edges between pixels in the same component cannot be cut. (c) We enrich the Multicut Problem by \emph{convexity constraints}. Our constraints implement the set theoretic definition of convexity on a discrete grid.
	}
\label{fig:method}
\end{figure*}

\textbf{Contributions.} In this work we propose the first model and solver for pixel-level segmentation of many generic convex objects of multiple foreground classes.
We introduce new models that include convexity constraints into \emph{multicut problems}, which are ILP formulations of image decomposition problems. We consider two multicut problems, one equivalent to the correlation clustering problem~\cite{bansal2004correlation}, another one equivalent to the Potts model. For efficiency, following the idea in~\cite{KappesEMMCVPR2011,AndresICCV2011}, we iteratively incorporate only the constraints that are violated per instance, and when no more violations occur, we are guaranteed the globally optimal solution. To the best of our knowledge, our models are the first to handle many convex objects and multiple foreground classes. Our models can be solved to global optimality in small yet practical cases. 
Figure~\ref{fig:method} gives an overview of the multicut problems as well as our proposed convexity constraints, as described in detail in Sections~\ref{sec:basics}~and~\ref{sec:method}. 


\section{Image Decomposition by Multicuts}
\label{sec:basics}
Given the pixel grid graph $G = (V,E)$ of an image, %
image segmentation tasks are often modeled as the problem of assigning, to each node $v\in V$, one label from a label set $L=\{1,\ldots,k\}$ so as to minimize some objective function. 
A widely used objective function is the energy of a pairwise conditional random field. A particular instance of this type of objective is the well-known Potts model. 
Its pairwise terms are non-zero only for edges that connect nodes with different labels. 
%
A special case of the Potts model is the \emph{Correlation Clustering} or \emph{Partitioning} model~\cite{bansal2004correlation}. The respective energy neglects unary terms, and the number of labels equals the number of nodes. 
%

In the following we describe two \emph{Multicut Problems}. The first one is equivalent to Correlation Clustering. The second one is equivalent to the Potts model. We also discuss respective solvers as proposed in~\cite{KappesEMMCVPR2011,AndresICCV2011}. 
These Multicut Problems are ILPs that form the basis of our contribution: In Section~\ref{sec:method} we take the set theoretic definition of convexity of the connected components of a segmentation, and directly translate it into inequality constraints that we add to the respective ILPs. 

\paragraph{The Minimum Cost Multicut Problem. } The minimum cost multicut problem~\cite{chopra1993partition} is equivalent to the correlation clustering problem. 
Its feasible solutions are all \emph{decompositions} of the graph $G$. 
A decomposition of $G$ is a partition $\Pi$ of $V$ such that, for every $U \in \Pi$, the subgraph of $G$ induced by $U$ is connected. 
A subgraph of $G$ that is connected and induced by a node set $U \subseteq V$ is called a \emph{component} of $G$. 
Any decomposition $\Pi$ of $G$ is characterized by the subset of edges that straddle distinct components: $E_\Pi = \{vw \in E \ | \ \forall U \in \Pi: v \notin U \vee w \notin U\}$.
Such a subset of edges is called a \emph{multicut} of $G$. There is exactly one multicut related to each decomposition of $G$. The following Theorem forms the basis of multicut problems: 
\begin{theorem}
The multicuts of $G$ are precisely the subsets $Y \subseteq E$ for which every cycle $C \subseteq E$ in $G$ satisfies $|C \cap Y | \neq 1$. 
\label{theo:1}
\end{theorem}
%
%
This has been proven in~\cite{chopra1993partition}. We give the proof in the Appendix. For a sketch, see Figure~\ref{fig:method}b. 

Let $y \in \{0,1\}^E$ be a 01-encoding of a multicut $Y$.
It makes explicit, for any pair $uv \in E$ of neighboring nodes, whether $u$ and $v$ are in distinct components, namely iff $y_{uv} = 1$.\footnote{Note that in~\cite{chopra1993partition} the interpretation of the 01-encoding is flipped, i.e.\ $y_e=0$ means that an edge is an element of the multicut.}
These \emph{edge labels} $y_e$ allow for an equivalent formulation of $|C \cap Y| \neq 1$ as linear inequality constraints~\eqref{eq:mc-cycles}, which leads to the following Definition: 
\begin{definition}
\label{def:multicut}
\cite{chopra1993partition}
Given a finite, simple, non-empty graph $G = (V, E)$ and a map $c: E \to \mathbb{R}$
(that is, for any pair $vw \in E$ of neighboring nodes, a cost or reward $c_{vw}$ for $v$ and $w$ being in distinct components),
the instance of the \emph{Minimum Cost Multicut Problem} (MC)
with respect to $G$ and $c$ 
is the ILP
\begin{align}
\min_{y \in \{0,1\}^E} \quad
    & \sum_{e \in E} c_e y_e 
        \label{eq:mc-objective}\\
\textnormal{subject to} \quad
    & \forall C \in \textnormal{cycles}(G)\ 
        \forall e \in C:\ 
            y_e \leq \sum_{\mathclap{e' \in C \setminus \{e\}}} y_{e'}
        \label{eq:mc-cycles}
\end{align}
\end{definition}
Constraints~\eqref{eq:mc-cycles} are referred to as as \emph{cycle constraints}. It is sufficient to consider only the \emph{chordless cycles} of $G$~\cite{chopra1993partition}.

\paragraph{The Minimum Cost Multicut Problem with Node Labels. } We also consider a Multicut Problem that is equivalent to the Potts model.  
This is a more constrained optimization problem in which every node assumes precisely one out of finitely many labels, and neighboring nodes are in the same component iff they have the same label:

\begin{definition}
\label{def:multiway-cut}
Given a finite, simple, non-empty graph $G = (V, E)$, 
a map $c: E \to \mathbb{R}$,
a finite set $L \neq \emptyset$
and a map $d: V \times L \to \mathbb{R}$
(that is, for any node $v$ and any label $l$, a cost or reward $d_{vl}$ for $v$ being labeled $l$),
the instance of the \emph{Minimum Cost Multicut Problem with Node Labels} (MCN)
with respect to $G$, $L$, $c$ and $d$
is the ILP
\begin{align}
\min_{\substack{x \in \{0,1\}^{V \times L}\\y \in \{0,1\}^E}} \quad
    & \sum_{e \in E} c_e y_e + \sum_{v \in V} \sum_{l \in L} d_{vl} x_{vl}
        \label{eq:lmc-objective}\\
\textnormal{subject to} \quad
    & \forall v \in V: \quad
        1 \leq \textstyle{\sum}_{l \in L} x_{vl}
        \label{eq:lmc-label-1}\\
    & \forall v \in V\ 
        \forall \{l,l'\} \in \tbinom{L}{2}: \quad
            x_{vl} + x_{vl'} \leq 1
        \label{eq:lmc-label-2}\\
    & \forall vw \in E\ 
        \forall \{l,l'\} \in \tbinom{L}{2}: \quad
            x_{vl} + x_{wl'} - 1 \leq y_{vw}
        \label{eq:lmc-cycle-1}\\
    & \forall vw \in E\
        \forall l \in L: \quad
            y_{vw} \leq 2 - x_{vl} - x_{wl}
        \label{eq:lmc-cycle-2}
\end{align}
\end{definition}

Here, any feasible solution $(x,y)$ is constrained such that every node $v$ is assigned at least and at most one label, namely the unique $l \in L$ such that $x_{vl} = 1$,
by \eqref{eq:lmc-label-1} and \eqref{eq:lmc-label-2}.
It is also constrained such that, for any edge $vw \in E$, $y_{vw} = 1$ if and only if $v$ and $w$ have distinct labels,
by \eqref{eq:lmc-cycle-1} and \eqref{eq:lmc-cycle-2}.
Thus, $y$ is the characteristic function of a multicut of $G$.
It defines uniquely a decomposition of $G$.

\paragraph{Solvers. } Branch-and-cut algorithms for Problem~\ref{def:multicut} are proposed in~\cite{KappesEMMCVPR2011,AndresICCV2011}. They find globally optimal solutions in reasonable run-time in many practical cases by including constraints~\eqref{eq:mc-cycles} per instance of the problem only in case they are violated. 
Problem~\ref{def:multiway-cut} is solved by~\cite{KappesEMMCVPR2011} by transforming it into an equivalent Minimum Cost Multicut Problem on a modified graph. This transformation involves flipping the meaning of node label variables resulting in the so-called \emph{multiway cut problem}. 

%
%
%
%
%
%
%
%
%



\section{Method}
\label{sec:method}
This Section provides the methodological contribution of our work. 
In Section~\ref{subsec:method:convexity-constraints-for-multicut}, we propose a model for image decomposition under the constraint that each component of the resulting partition has to be convex. We achieve this via additional inequality constraints that we include into the Minimum Cost Multicut Problem~\ref{def:multicut}. 
Furthermore, we propose a respective model with node labels. 
With this model, we can enforce for any pair of labels, $\{k,l\} \in \tbinom{L}{2}$, that a component of label $k$ does not contain any node labeled $l$ in its convex hull. 
This is more general than ``simply'' enforcing convexity of components. 
We achieve this, again, via inequality constraints that we include into the Minimum Cost Multicut Problem with Node Labels~\ref{def:multiway-cut}. 
In Section~\ref{subsec:method:optimization} we propose a solver for the above optimization problems.


\subsection{Convexity Constraints for Image Decomposition}
\label{subsec:method:convexity-constraints-for-multicut}
\begin{figure*}
\begin{center}
   \includegraphics[width=0.92\textwidth]
                   {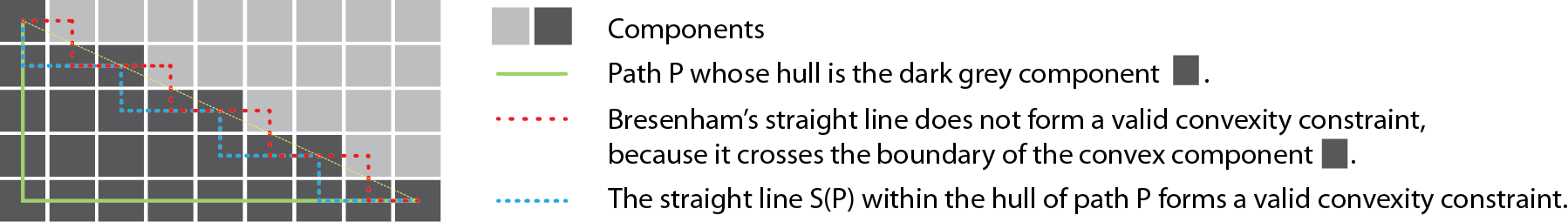} 
\end{center}
   \caption{Discrete convexity in image grid graphs. 
	}
\label{fig:straightlines}
\end{figure*}

Let $P\subset E$ denote an arbitrary open path in $G=(V,E)$. 
All components of an image decomposition $\Pi$ are convex iff for any path $P$ that does not contain any edges in $E_\Pi$, 
the straight line between the end points of $P$ also does not contain any edges in $E_\Pi$. 
We discretize this set theoretic definition of convexity as follows: 
In a pixel grid graph with the usual embedding into the 2d plane, where pixels are Voronoi regions of graph nodes (cf.\ Fig.\ \ref{fig:method}a), 
a component is \emph{discrete convex} iff for every path $P$ that does not contain any edges in $E_\Pi$, the interior of the loop formed by $P$ and the straight line between its end points does not enclose any nodes of a distinct component. See Figure~\ref{fig:straightlines} for a sketch.
We call the set of nodes enclosed by this loop the \emph{hull} of $P$.
Let $S(P)$ denote the path in $G$ that runs along the boundary of the discrete hull of $P$ and connects the first and the last node covered by $P$ (cf.\ Fig.\ \ref{fig:straightlines}). 
All components of an image partition are discrete convex iff 
\begin{equation}
\forall P\in \textnormal{paths}(G): \quad \sum\limits_{e\in P} y_{e} = 0 \Rightarrow \sum\limits_{e\in S(P)} y_{e} = 0 .
\label{eq:convexity-constraints-as-implication}
\end{equation}
For a sketch, see Figure~\ref{fig:method}c. 
Note that in general, the Bresenham line~\cite{bresenham1965algorithm} is different from $S(P)$, as sketched in Figure~\ref{fig:straightlines}.
We propose to formulate \eqref{eq:convexity-constraints-as-implication} as linear inequality constraints, 
which enables us to formulate the task of finding the optimal decomposition of $G$ into convex components as an ILP:

\begin{definition}
\label{def:multicut-convex}
Given a finite, simple, non-empty graph $G = (V, E)$ and a map $c: E \to \mathbb{R}$,
the instance of the \emph{Minimum Cost Convex Component Multicut Problem} (Convex-MC)
with respect to $G$ and $c$ 
is the ILP
\begin{align}
\min_{y \in \{0,1\}^E} \quad
    & \sum_{e \in E} c_e y_e \quad 
 \textnormal{subject to} \quad \eqref{eq:mc-cycles} \\
 \textnormal{and} \quad &
     \forall P \in \textnormal{paths}(G): \quad |S(P)| \cdot \sum\limits_{e\in P} y_{e} \geq \sum\limits_{e \in S(P)} y_{e} 
        \label{eq:mc-convexity}
\end{align}
\end{definition}
We also refer to this model as \emph{correlation clustering with convexity constraints}, to put it into well-known terminology.
\begin{lemma} 
\label{lemma:1}
Constraints~\eqref{eq:mc-convexity} are equivalent to \eqref{eq:convexity-constraints-as-implication}.
\end{lemma} 
A proof of Lemma~\ref{lemma:1} is given in the Appendix. 


Now we consider image decomposition with node labels, which is equivalent to the Potts model (cf.\ Section~\ref{sec:basics}). 
%
Let $V_P$ denote the nodes covered by a path $P$, and $V_{S(P)}$ the nodes covered by the respective straight line $S(P)$ but not by $P$ (i.e., all but the ``end points'' of $S(P)$). 
Given an image decomposition with node labels $x$, all components assigned to label $k$ are convex iff
\begin{equation}
\forall P: \sum\limits_{v\in V_P} x_{vk} = | V_P | \Rightarrow \sum\limits_{v\in V_{S(P)}} x_{vk} = |V_{S(P)}| .
\label{eq:convexity-constraints-multiway-as-implication}
\end{equation}
%
%
More generally, components assigned to label $k$ have in their convex hulls only nodes with labels from a subset of $L_k \subset L$ (where $k\in L_k$) iff
\begin{equation}
\begin{split}
\forall P: & \ \sum\limits_{v\in V_P} x_{vk} = | V_P|   
               \ \Rightarrow \sum\limits_{\substack{v\in V_{S(P)}, \\ l \in L_k} } x_{vl} = |V_{S(P)}| .
\label{eq:convexity-constraints-relative-multiway-as-implication}
\end{split}
\end{equation}
This holds because~\eqref{eq:lmc-label-2} entails $x_{vk} = 1 \Rightarrow \sum_{l\in L_k} x_{vl} = 1$. 
Constraints \eqref{eq:convexity-constraints-multiway-as-implication} are a special case of \eqref{eq:convexity-constraints-relative-multiway-as-implication}, namely with $L_k = \{k\}$. 
We propose to formulate~\eqref{eq:convexity-constraints-relative-multiway-as-implication} as linear inequality constraints and hence yield an ILP that models image decomposition, with node labels, into convex components:

\begin{definition}
\label{def:multiway-cut-convex}
Given a finite, simple, non-empty graph $G = (V, E)$, 
a map $c: E \to \mathbb{R}$,
a finite set $L \neq \emptyset$,
a map $d: V \times L \to \mathbb{R}$,
and a set $L_k\subset L$ for each $k\in L$ with $k\in L_k,$
the instance of the \emph{Minimum Cost Convex Component Multicut Problem with Node Labels} (Convex-MCN)
with respect to $G$, $L$, $c$, $d$, and $\{L_k:k\in L\}$
is the ILP
\begin{align}
\min_{\substack{x \in \{0,1\}^{V \times L}\\y \in \{0,1\}^E}} \quad
    & \sum_{e \in E} c_e y_e + \sum_{v \in V} \sum_{l \in L} d_{vl} x_{vl}
        \label{eq:convex-lmc-objective}
\quad \textnormal{subject to} \quad
    \eqref{eq:lmc-label-1}, \eqref{eq:lmc-label-2}, \eqref{eq:lmc-cycle-1}, \eqref{eq:lmc-cycle-2}\\
  \textnormal{and} \quad  & \forall P: \quad 
        |V_{S(P)}| \cdot \left( |V_P| - \sum\limits_{v\in V_P} x_{vk} \right)
  \geq |V_{S(P)}| - \sum\limits_{\substack{v\in V_{S(P)}, \\ l \in L_k}} x_{vl} 
  \label{eq:convexity-ilp-constraints-relative-multiway}
\end{align}
\end{definition}
We also refer to this model as \emph{Potts model with convexity constraints}, to put it into well-known terminology.
\textcolor{red}{The proof of equivalence between the convexity constraints \eqref{eq:convexity-constraints-relative-multiway-as-implication} and \eqref{eq:convexity-ilp-constraints-relative-multiway} works analogously to the proof of Lemma~\ref{lemma:1} that we give in the Appendix. }


\subsection{Optimization} 
\label{subsec:method:optimization}

\begin{figure*}[h]
\begin{center}
   \includegraphics[width=0.8\textwidth]
                   {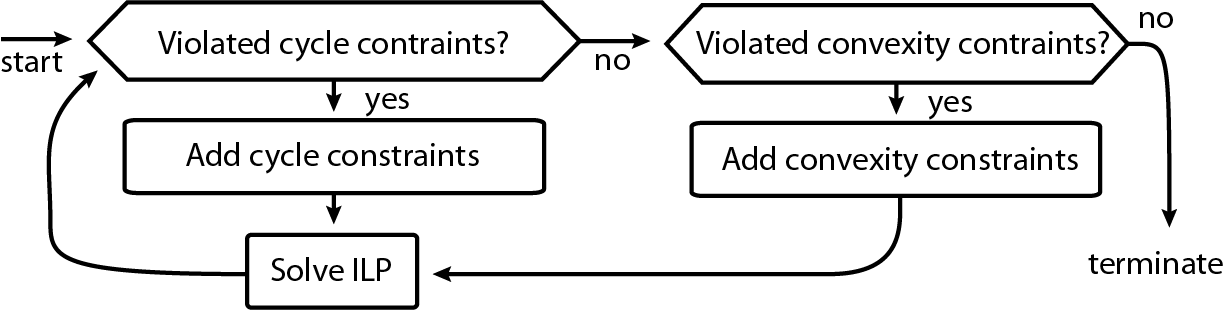} 
\end{center}
   \caption{Our proposed solver for our Convex Component Multicut Problems (Definitions~\ref{def:multicut-convex} and~\ref{def:multiway-cut-convex}).}  
\label{fig:algorithm}
\end{figure*}

Similar to~\cite{KappesEMMCVPR2011,AndresICCV2011} we pursue a cutting plane approach in which violated cycle- and convexity constraints are separated and added incrementally. 
Our algorithm starts by solving MC (see Def.\ \ref{def:multicut}) or MCN (see Def.\ \ref{def:multiway-cut}) with the method described in~\cite{KappesEMMCVPR2011}. 
Given the resulting image decomposition, we solve the Separation Problem w.r.t.\ convexity constraints as described below.
%
%
However, solving an ILP with added convexity constraints can entail new violations of \textcolor{red}{cycle constraints. }
Hence we propose the algorithm defined in Figure~\ref{fig:algorithm}.
%
%
Upon termination, the algorithm gives the globally optimal feasible solution. 
In practice we can, optionally, allow for ILPs to be solved up to some relative gap of $\epsilon \%$. In this case our algorithm gives a feasible solution whose energy is at most $\epsilon_{last} \%$ away from the global optimum, where $\epsilon_{last} \%$ is the gap obtained in the last iteration, i.e.\ directly before termination. 


\textbf{Separation of Convexity Constraints. }
%
%
Given an image decomposition, assuming a 2D pixel grid, we detect violated convexity constraints by scanning along sequences of nodes, $\{v_1, \ldots, v_n \}$, 
each defined by a start node $v_1 = (x,y)$ and an offset $(a,b)$, with $a,b \in \mathbb{N}$ co-prime. 
In case of the Convex-MC problem~\ref{def:multicut-convex}, convexity is violated iff two nodes $v_i, v_j$ along a sequence lie in the same component, and there exists a node $v_z$ in the sequence with $i<z<j$ such that $v_z$ lies in a different component. 
In case of the Convex-MCN problem \ref{def:multiway-cut-convex}, there is a violation of (generalized) convexity iff $v_i, v_j$ lie in the same component, labeled $k$, and there exists a node $v_z$ with $i<z<j$ such that $v_z$ is labeled $l\not\in L_k$. 

The cardinality of the set of orientations of sequences, $\mathcal{A}:=\{(a,b) \ \textnormal{co-prime,} \ -N_x\leq a \leq N_x, N_y\leq b \leq N_y\}$, is \textcolor{red}{bounded by $O(N^2)$,} where $N=N_x\cdot N_y$ denotes the number of pixels in a 2d image. 
Scanning for violations along a particular sequence can be done greedily. For each orientation, the union of the sets of sequences to consider covers the whole image. 
Hence the computational complexity for processing one particular orientation is in $O(N)$. 
Overall, the computational complexity of scanning for violations is \textcolor{red}{bounded by $O(N^3)$. }
Optionally, one may want to consider only a subset of orientations, $A\subset\mathcal{A}$. Then the effort reduces to $O(|A|\cdot N)$.
\label{subsec:method:paths}
Given a pair of nodes $(v_i,v_j)$ that constitutes a violation of convexity, we need to find a respective path $P$ and straight line $S(P)$. 
%
As for a path $P$ from $v_i$ to $v_j$ through their component: 
In general there are many such paths.
To form a constraint, we pick just one such path.
We explore two variants:
In case of Convex-MC (see Def.\ \ref{def:multicut-convex}), a shortest path in terms of number of edges;
In case of Convex-MCN (see Def.\ \ref{def:multiway-cut-convex}), a cheapest path in terms of unaries of the label assigned to the component. 
Dijkstra's algorithm yields an optimal path for either variant. 

%
%
\textcolor{red}{
We compute the straight line $S(P)$ as follows:
(1) Determine the orientation of the loop formed by path P and the (continuous) straight line $v_i,v_j$. 
(2) Infer the direction of the normal on the (continuous) straight line $v_i,v_j$ that points to the interior of this loop. Without loss of generality, we assume that $P$ and the continuous straight line do not intersect. 
(3) Greedily find the discrete shortest path with minimal distance of nodes to the continuous straight line, subject to the constraint that no node lies outside the loop. 
}




\section{Results and Discussion}

We present proof-of-concept results of our method on various photographs and biological images. For each exemplary image, in addition to the result of our method, we also show the respective result obtained without convexity constraints. Finally, we provide a comparison to state of the art~\cite{BoykovECCV2014} on two exemplary images. 
We employ a four-connected grid graph in all experiments. 
If not noted otherwise, we check violations of convexity in 8 discrete directions, 
and set the stopping criterion for the ILP solver to 2\%. 
This is not a hard stopping criterion, and hence smaller gaps are achieved per instance. 
Table~\ref{tab:runtimes_energies} lists the gaps, number of iterations, run-times, and energies obtained for each of the examples. 


%
\label{results:subsec:proof-of-concept} 
\begin{figure*}[p]
\setlength\tabcolsep{1pt}
\begin{center}
\begin{tabular}{cccccc}

   \includegraphics[width=0.161\textwidth]
                   {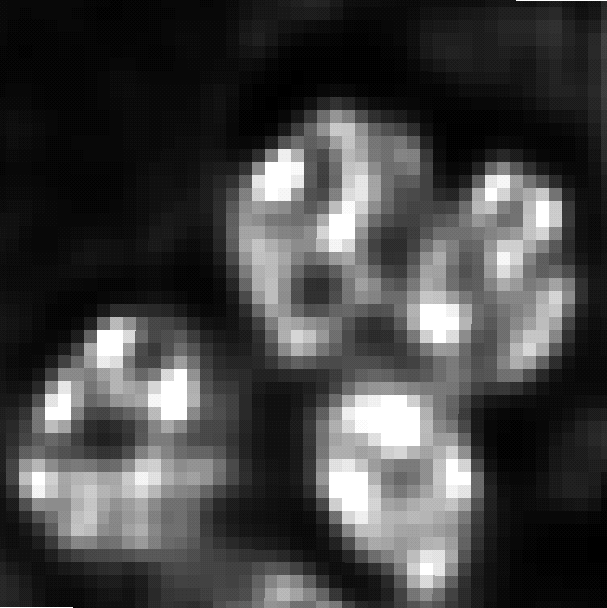} &
   \includegraphics[width=0.161\textwidth]
                   {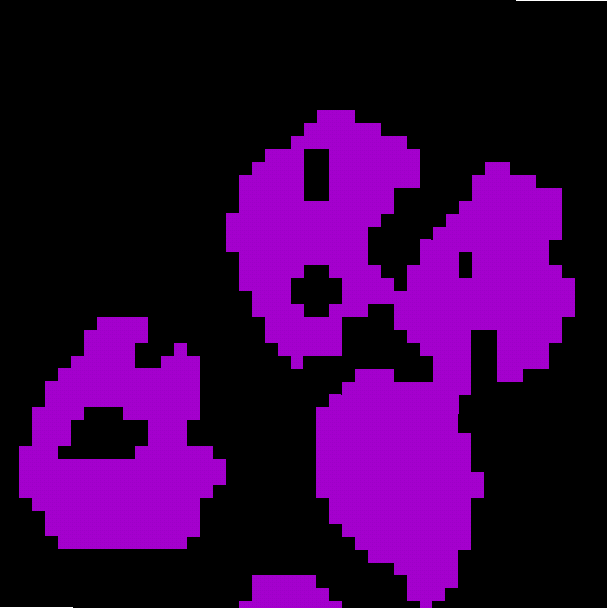} & 
   \includegraphics[width=0.161\textwidth]
                   {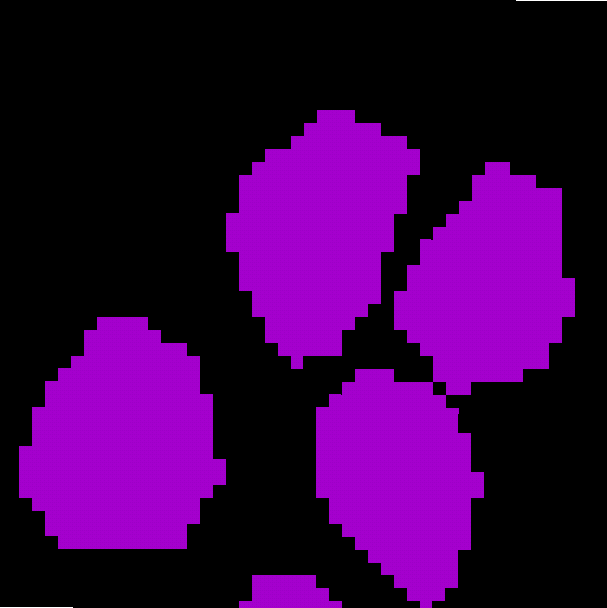} &
   \includegraphics[width=0.161\textwidth]
                   {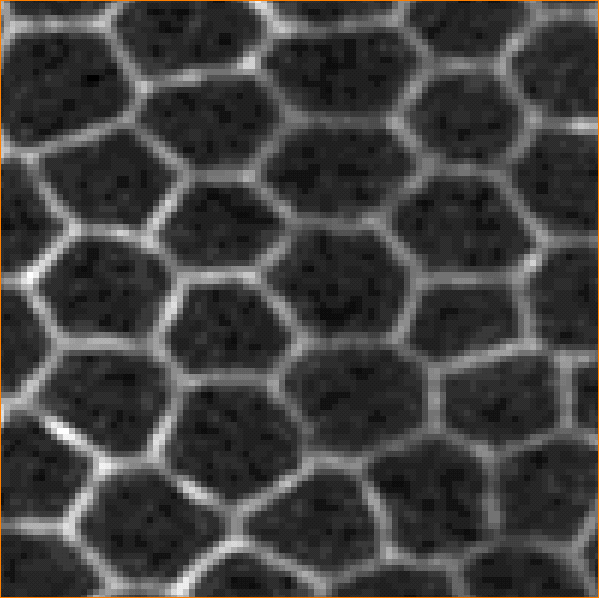} &
   \includegraphics[width=0.161\textwidth]
                   {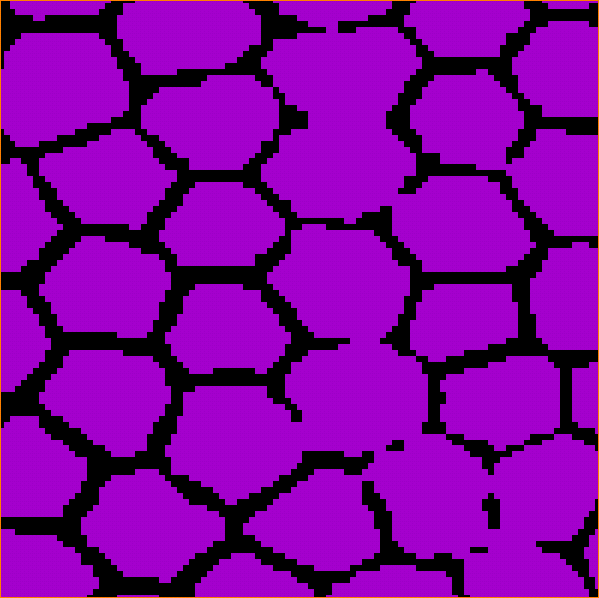} & 
   \includegraphics[width=0.161\textwidth]
                   {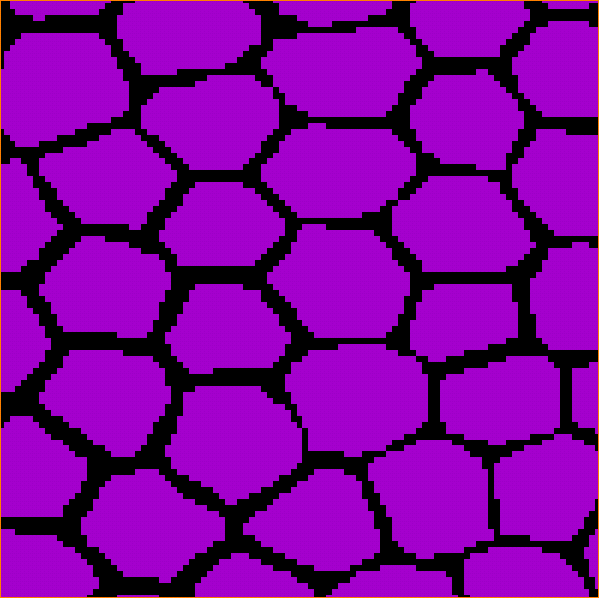} \\
									(a) & (b) & (c) & (d) & (e) & (f) 
\vspace{-5pt}
\end{tabular}
\end{center}
   \caption{ Examples for convexity constraints in a two-label Potts model: Excerpts of microscopic images of (a)~cell nuclei of a nematode worm, and (d) densely packed cells in a fly wing. (b/e)~Two-label Potts model. (c/f)~Convexity constraints on foreground label of the respective Potts model. Note that for the nuclei we check violations of convexity in \emph{all} discrete directions. 
	}
\label{fig:2LabelPotts}
\end{figure*}
\begin{figure*}[p]
\setlength\tabcolsep{1pt}
\begin{center}
\begin{tabular}{cccccc}

   \includegraphics[width=0.161\textwidth]
                   {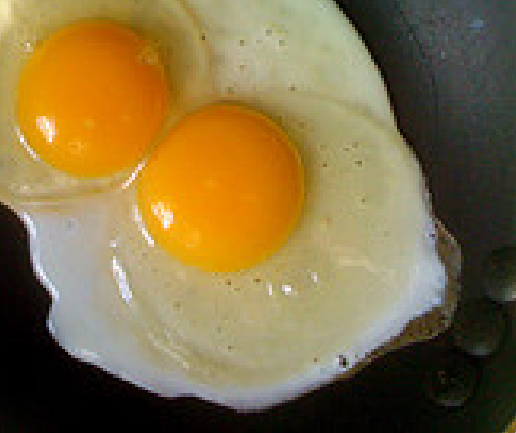} &
   \includegraphics[width=0.161\textwidth]
                   {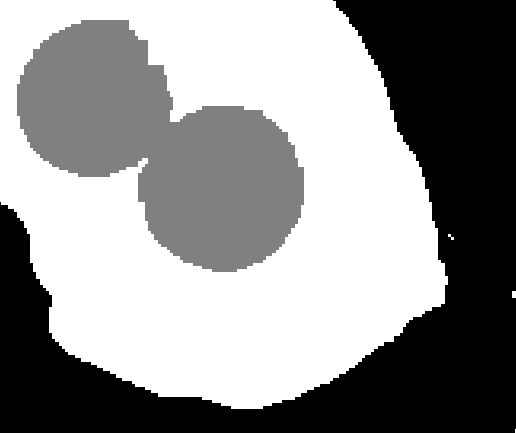} & 
   \includegraphics[width=0.161\textwidth]
                   {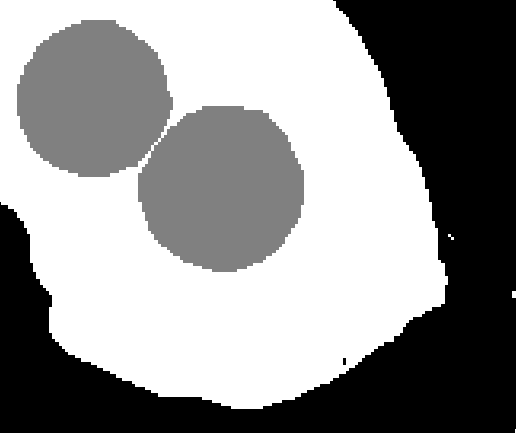} &
									  %
   \includegraphics[width=0.161\textwidth]
                   {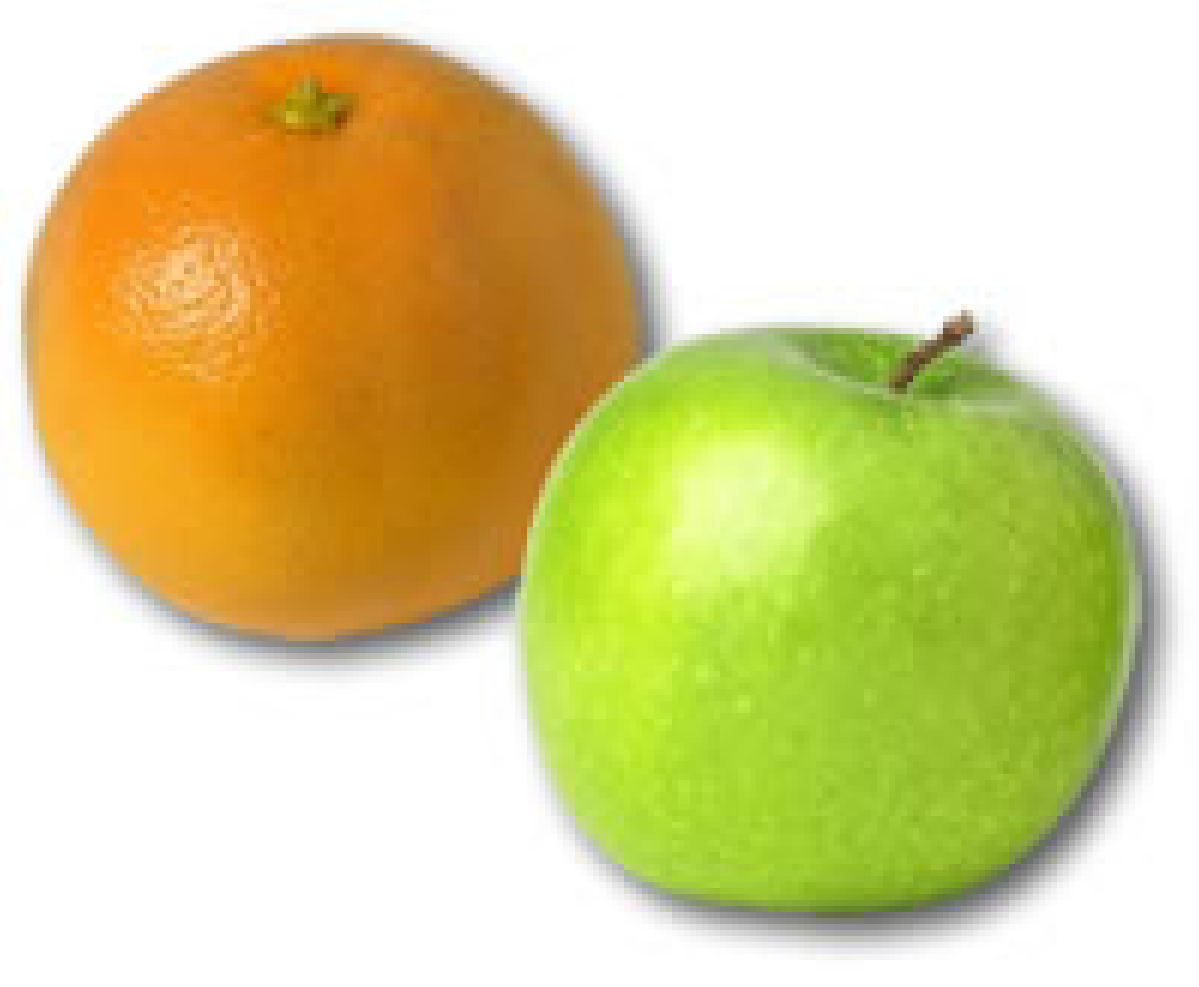} &
   \includegraphics[width=0.161\textwidth]
                   {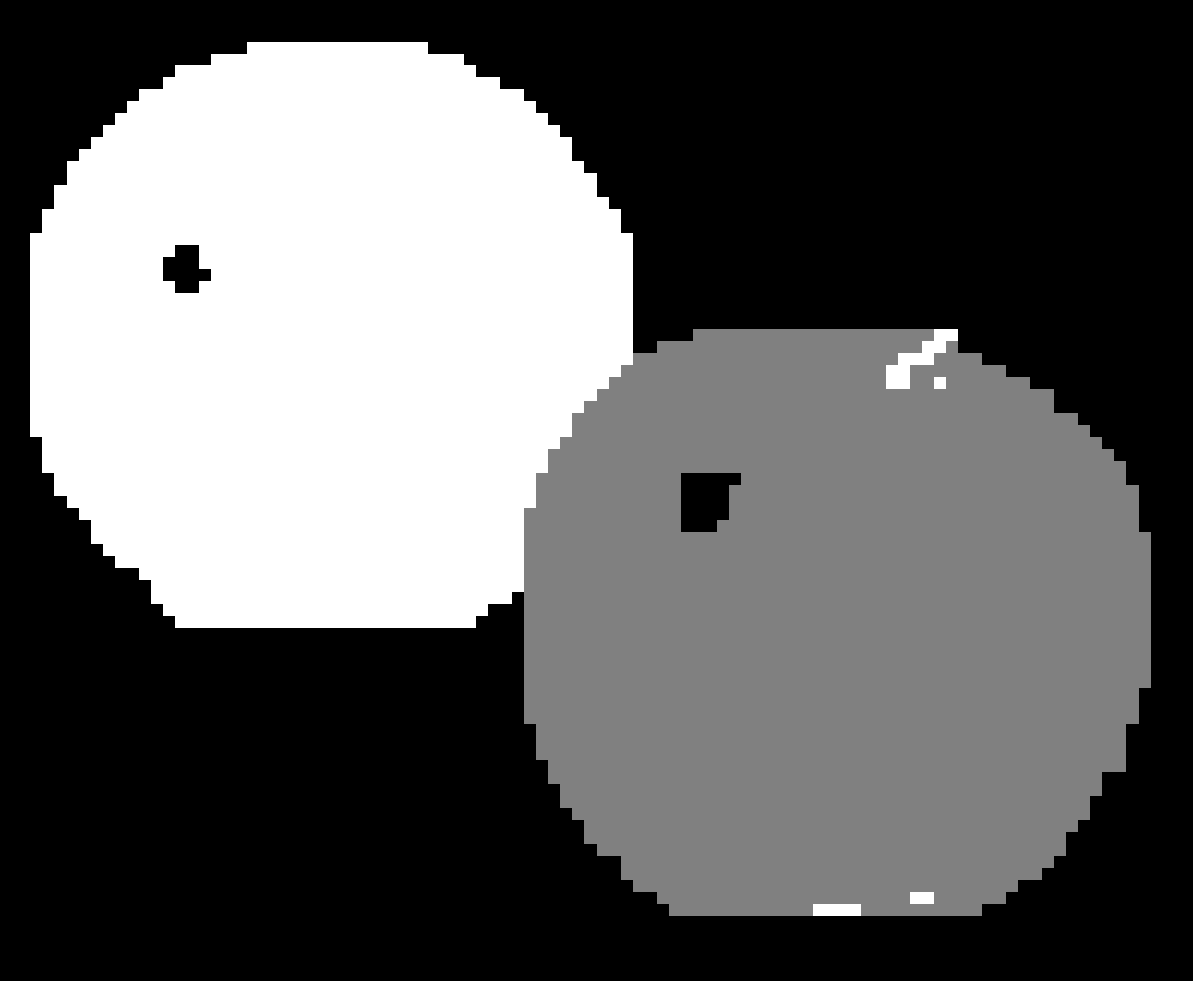} & 
   \includegraphics[width=0.161\textwidth]
                   {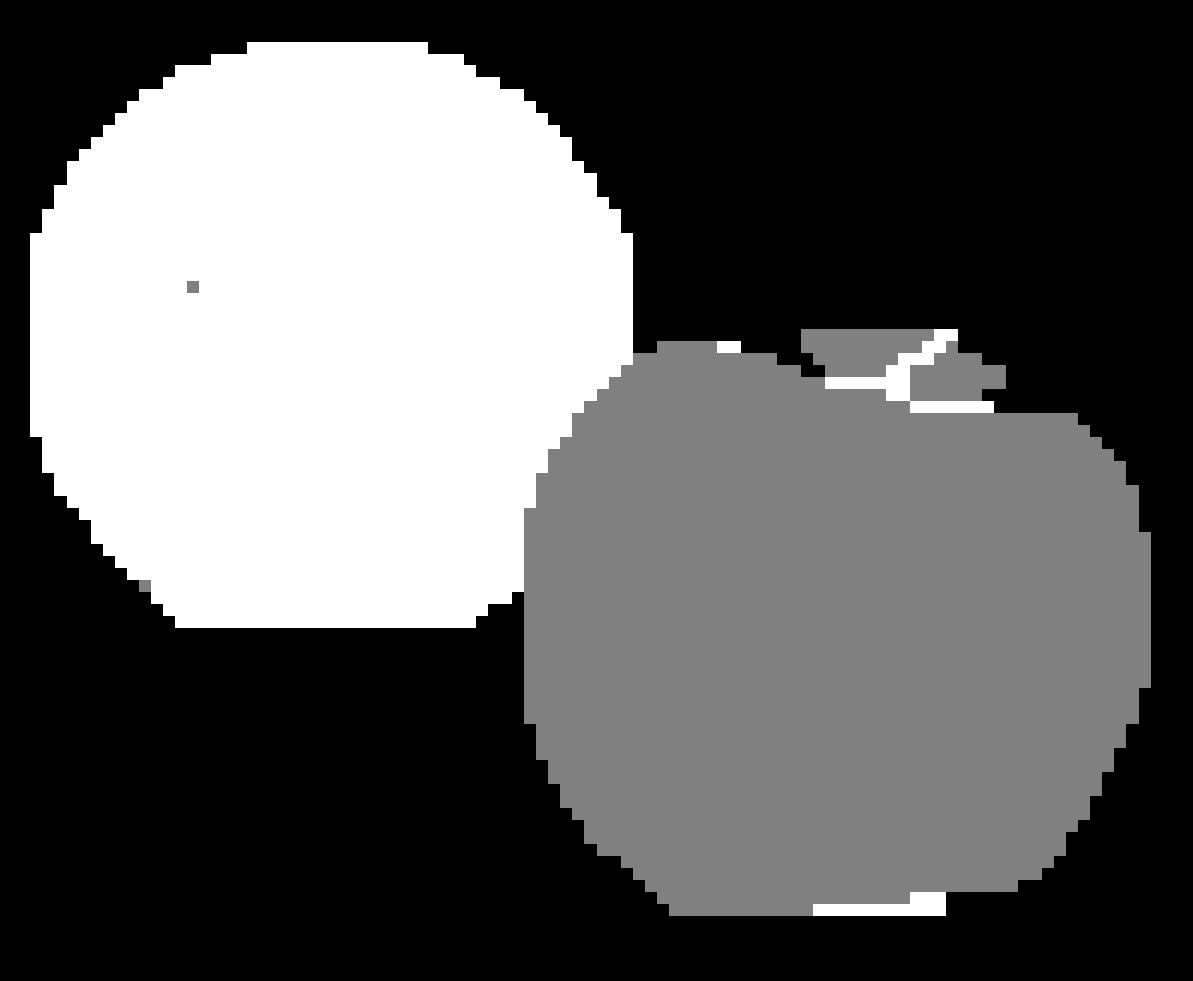} \\
									 (a) & (b) & (c) & (d) & (e) & (f)
\vspace{-5pt}
\end{tabular}
\end{center}
   \caption{ Examples for convexity constraints in three-label Potts models. (a)~\href{https://flic.kr/p/96wjK6}{Photograph} of two fried eggs in a pan (by Matthew Hurst on flickr, \href{https://creativecommons.org/licenses/by-sa/2.0/}{CC BY-SA 2.0}). (b)~Three-label Potts model. (c)~Convexity constraints on yolk  label of the same three-label Potts model. 
(d)~\href{http://www.ohioindividualhealthinsurance.net/wp-content/uploads/2013/06/apple-orange.jpg}{Image} of an apple occluding an orange. (e)~Three-label Potts model. (f)~Generalized convexity constraints, with $L_{apple} = \{apple\}$ and $L_{orange} = \{apple,orange\}$. Thus the apple label cannot have anything but apple in it's convex hull, while the orange label is allowed to have apple in it's convex hull, but not background. 
}
\label{fig:3LabelPotts}
\label{fig:apple-orange}
\end{figure*}
\begin{figure*}[p]
\setlength\tabcolsep{1pt}
\begin{center}
\begin{tabular}{cccccc}

   \includegraphics[width=0.161\textwidth]
                   {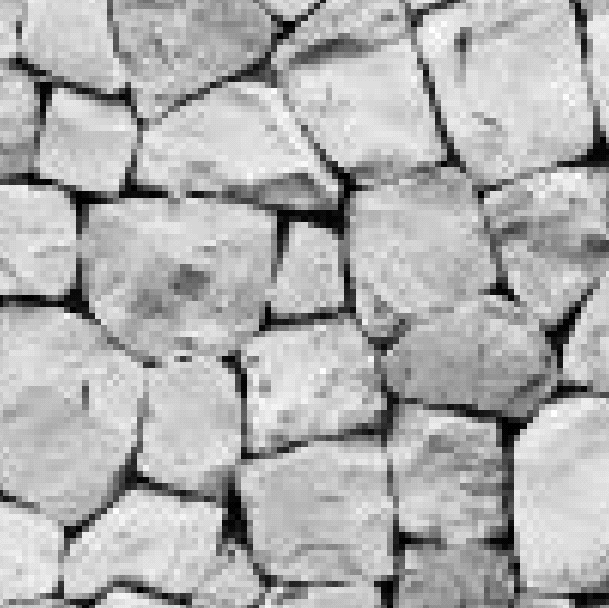} &
   \includegraphics[width=0.161\textwidth]
                   {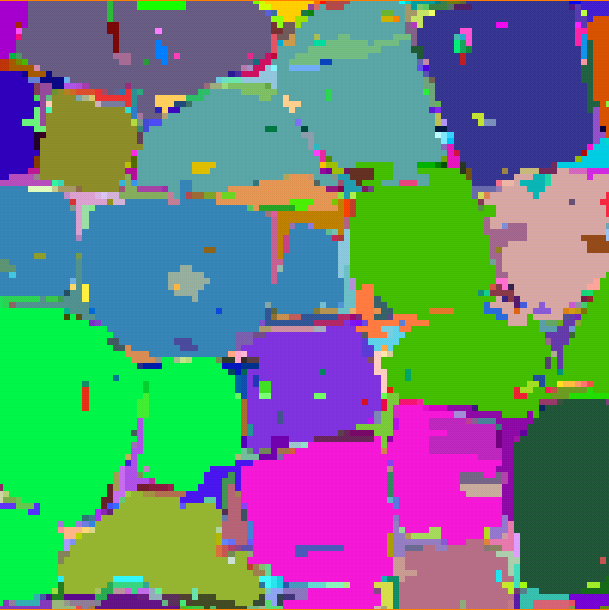} & 
   \includegraphics[width=0.161\textwidth]
                   {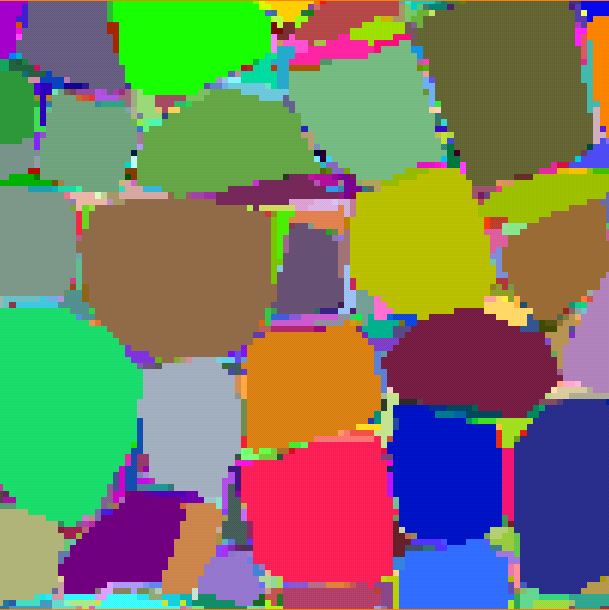} &
   \includegraphics[width=0.161\textwidth]
                   {gfx/flywing-image.png} &
	 \includegraphics[width=0.161\textwidth]
                   {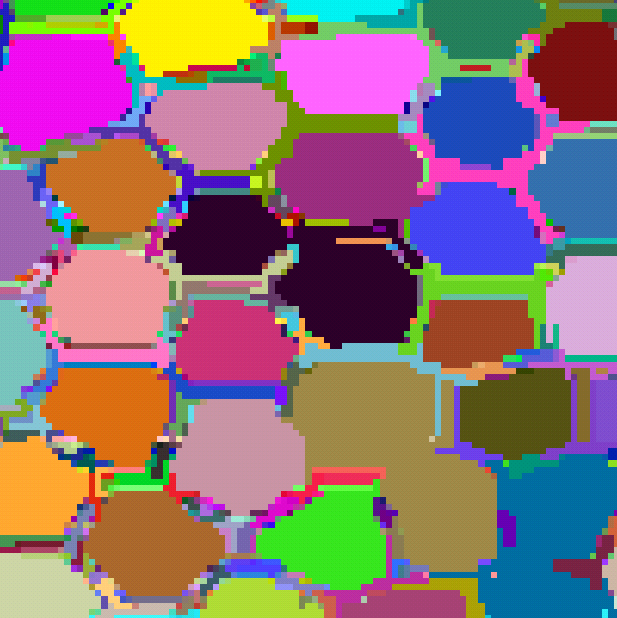} & 
	 \includegraphics[width=0.161\textwidth]
                   {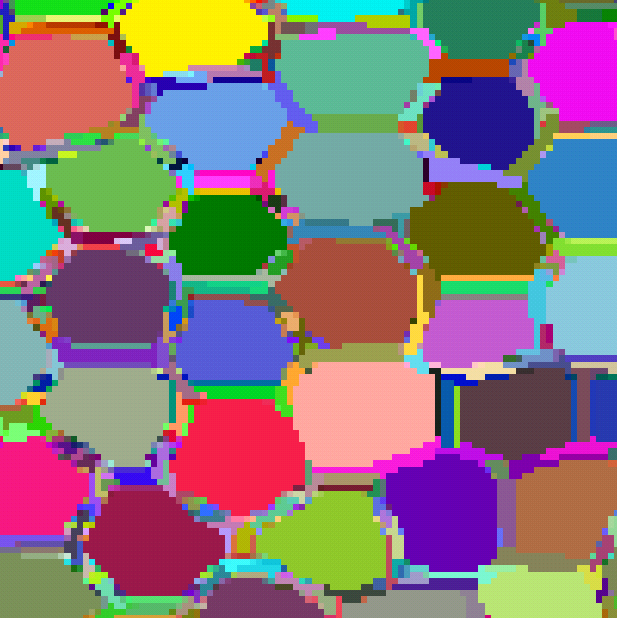} \\
									(a) & (b) & (c) & (d) & (e) & (f)
\vspace{-5pt}
\end{tabular}
\end{center}
   \caption{ Examples for convexity constraints in correlation clustering models: (a) Natural image of a stone wall. (d) Densely packed cells in excerpt of microscopic image of fly wing. (Same image as in Figure~\ref{fig:2LabelPotts}(d).) (b/e) Correlation clustering model. (c/f) Convexity constraints in the respective correlation clustering model. }
\label{fig:CorrelationClustering}
\end{figure*}
\begin{table*}[p]%
\small
\begin{center}
\begin{tabular}{lcrrrrr}
Experiment & Res & E & ConvexE & \# Iter & Time & Gap\\
\hline
Nuclei, 2 label Potts	 		         & 48x48   & 725   & 770   & 9  & 15 sec  & 0.05 \% \\
Fly wing, 2 label Potts 		       & 101x101 & 1089  & 1240  & 11 & 380 sec & 2 \% \\
Fried eggs, 3 label Potts	         & 163x137 & 64633 & 64567 & 22 & 166 sec & 0.04 \% \\
Synthetic, 3 label Potts           & 155x104 & 3275  & 3529  & 1  & 2 sec   & 0 \% \\
Apple and orange, 3 label Potts    & 200x165 & 40390 & 40428 & 158 & 870 sec & 0.06 \% \\
Stone wall, correlation clustering & 101x101 & -226  & -188  & 33 & 10519 sec & 1.5 \% \\
Fly wing, correlation clustering   & 101x101 & -419  & -394  & 14 & 251 sec & 1 \% \\
\hline 
\vspace{0pt}
\end{tabular}
\caption{For each experiment, we list the image resolution in pixels (Res), the energy of the solution obtained without convexity constraints (E), the energy of the solution obtained with convexity constraints (ConvexE), the number of times that convexity constraints are iteratively added to the ILP (\# Iter), the run-time of the algorithm with convexity constraints (Time), and the gap achieved in the final iteration (Gap). }
\label{tab:runtimes_energies}
\end{center}
\end{table*}
\begin{figure*}
\setlength\tabcolsep{1pt}
\begin{center}
\begin{tabular}{ccccc}

   \includegraphics[width=0.165\textwidth]
                   {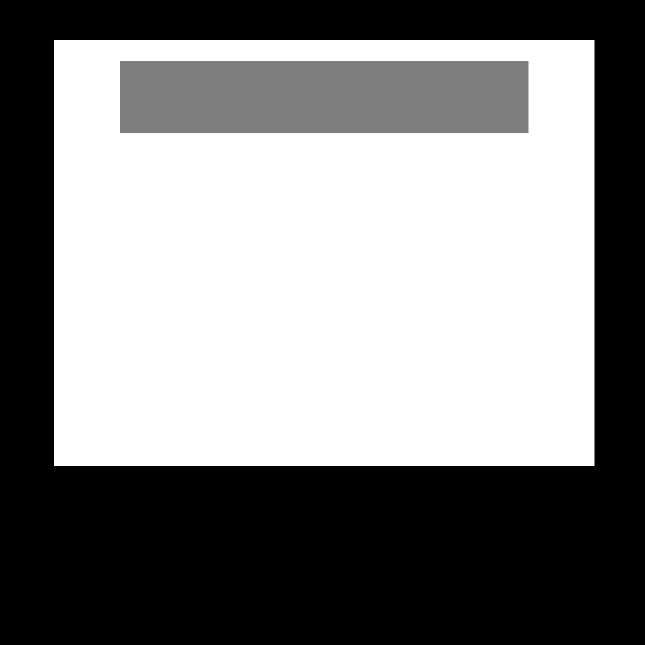} &
   \includegraphics[width=0.165\textwidth]
                   {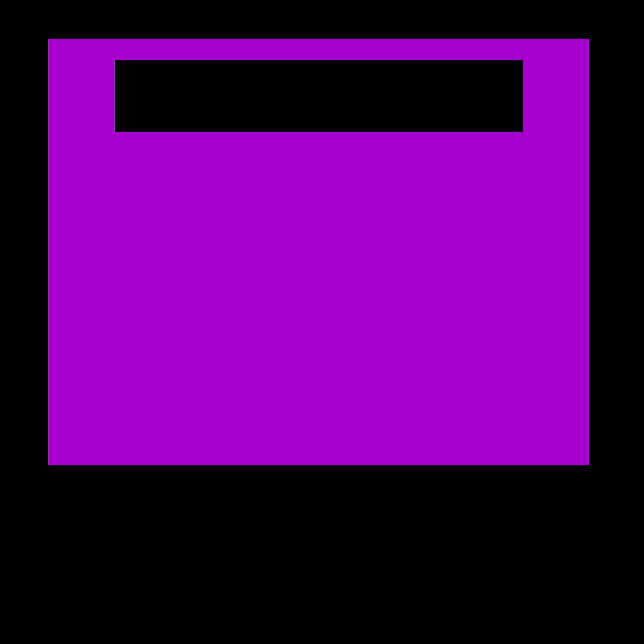} & 
   \includegraphics[width=0.165\textwidth]
                   {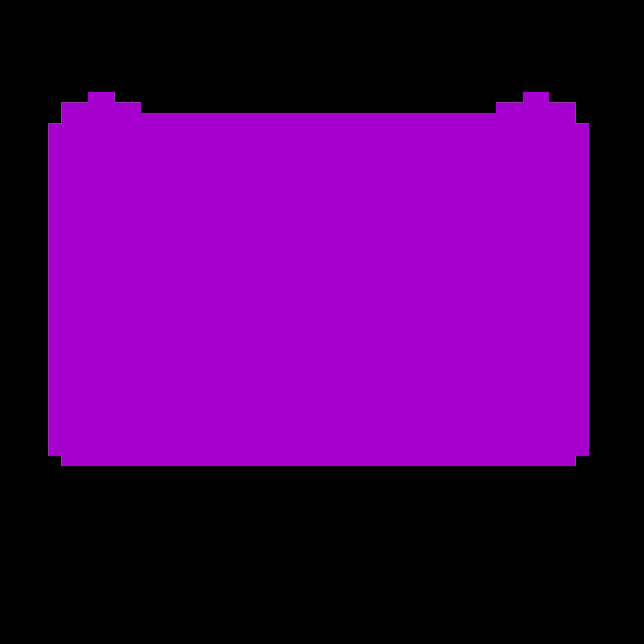} &
	 \includegraphics[width=0.165\textwidth]
                   {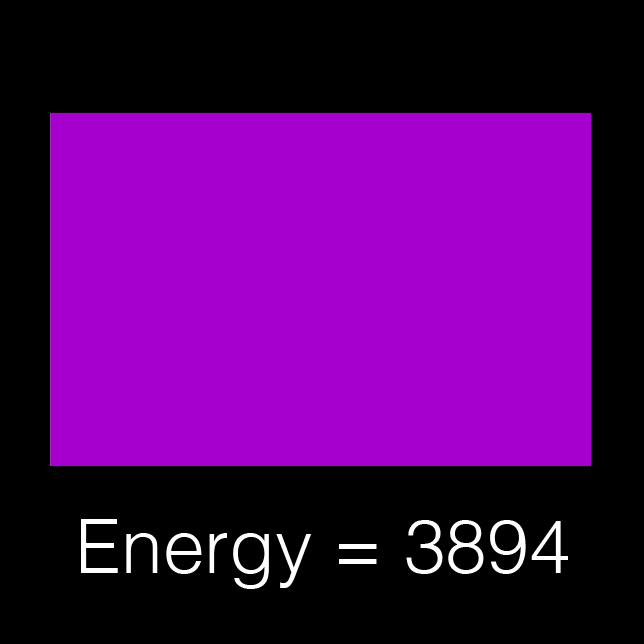} &
                      \includegraphics[width=0.165\textwidth]
                   {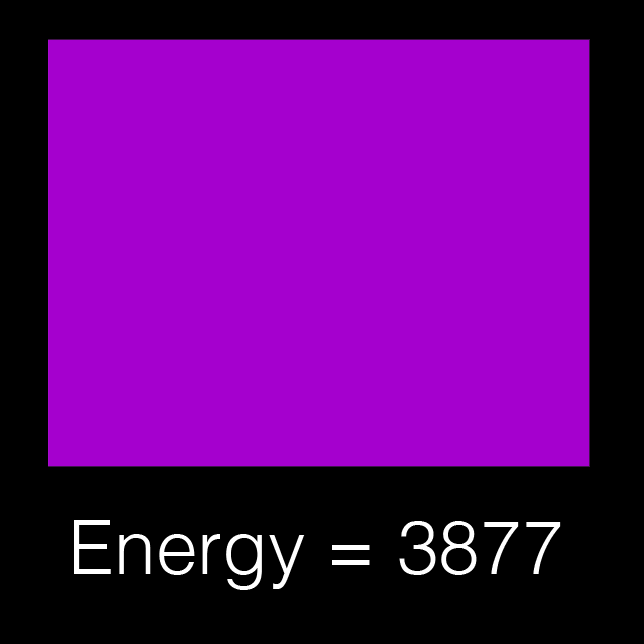} \\
   \includegraphics[width=0.15\textwidth]
                   {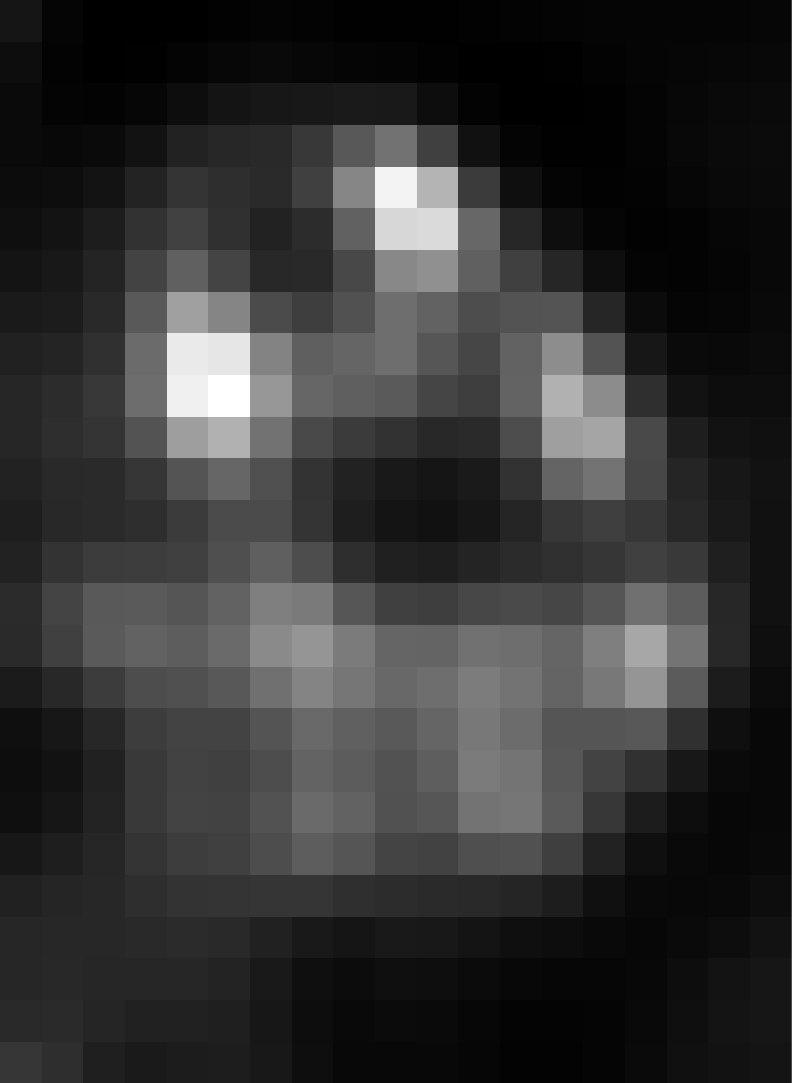} &
   \includegraphics[width=0.15\textwidth]
                   {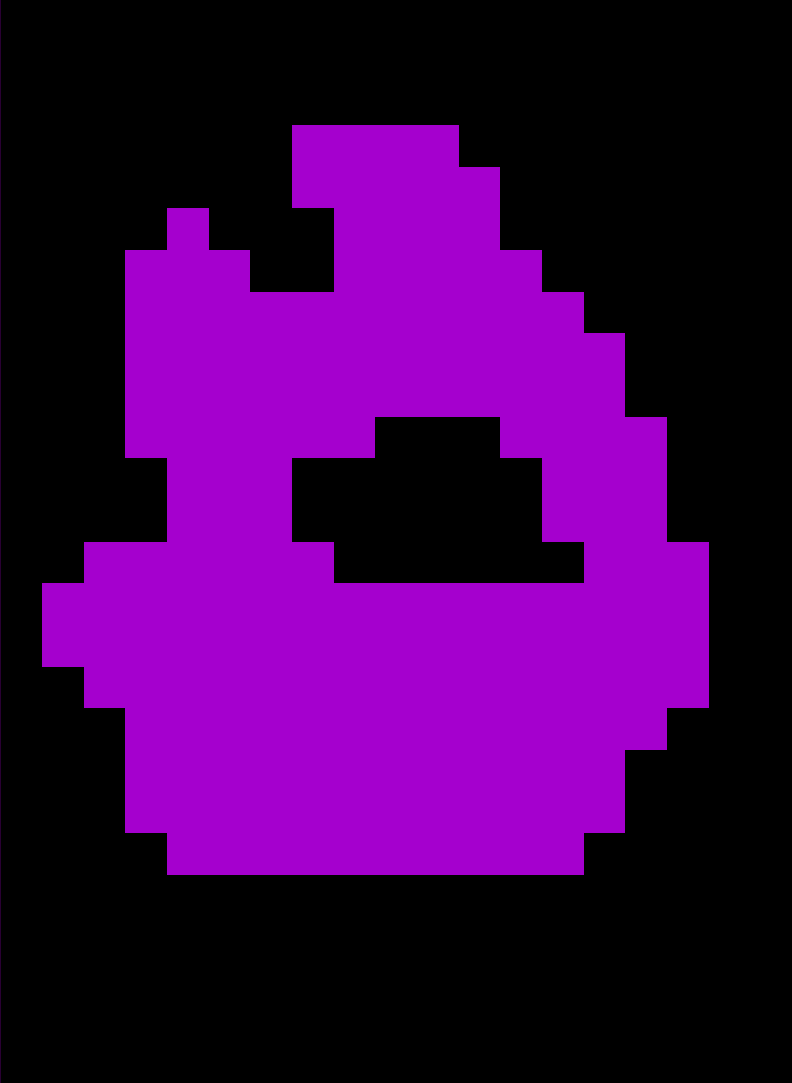} & 
   \includegraphics[width=0.15\textwidth]
                   {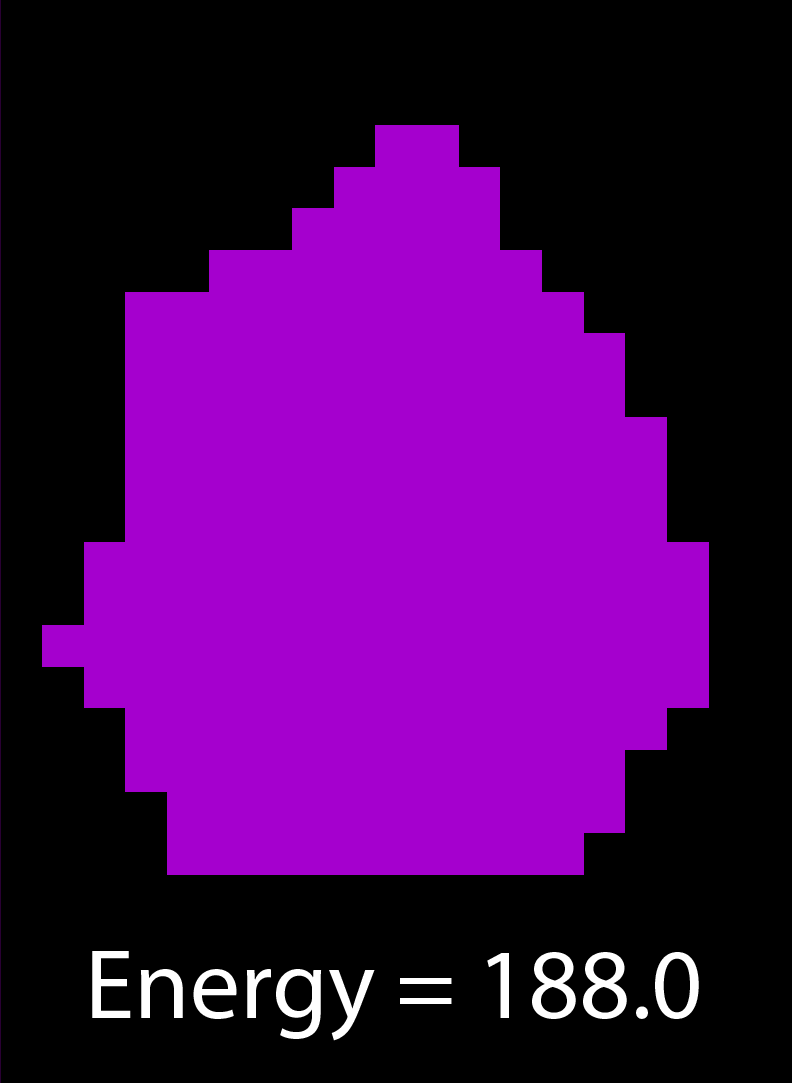} &
	 \includegraphics[width=0.15\textwidth]
                   {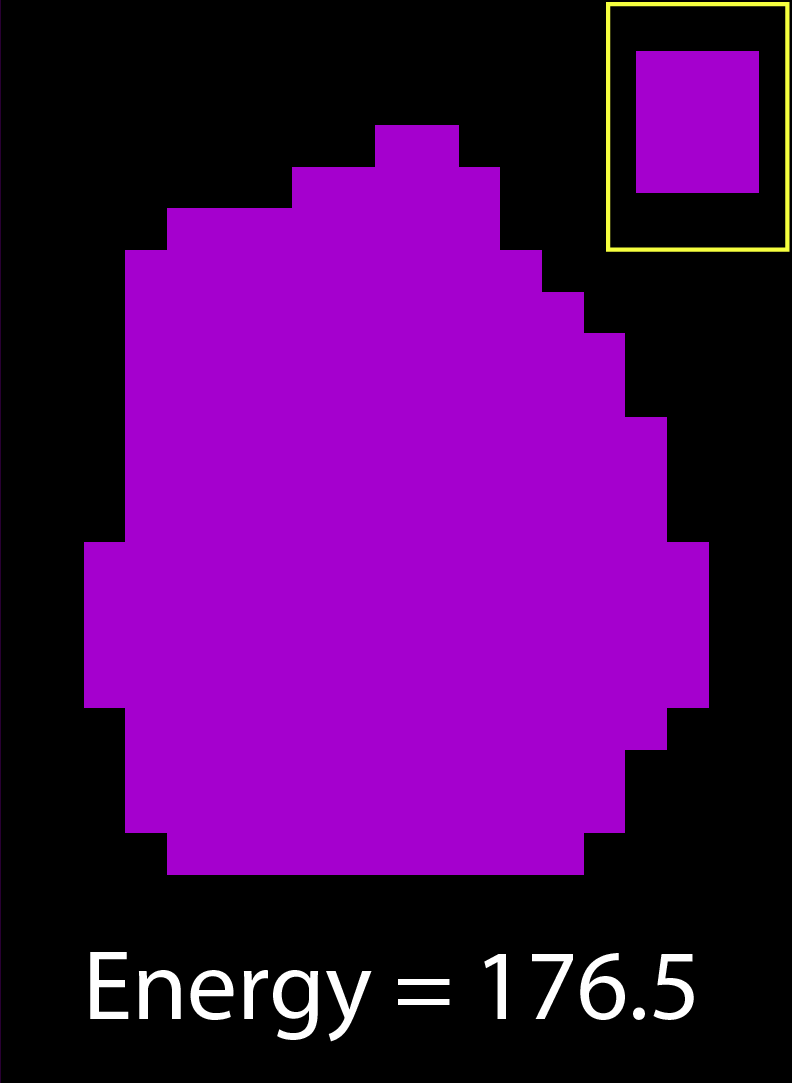} &
                      \includegraphics[width=0.15\textwidth]
                   {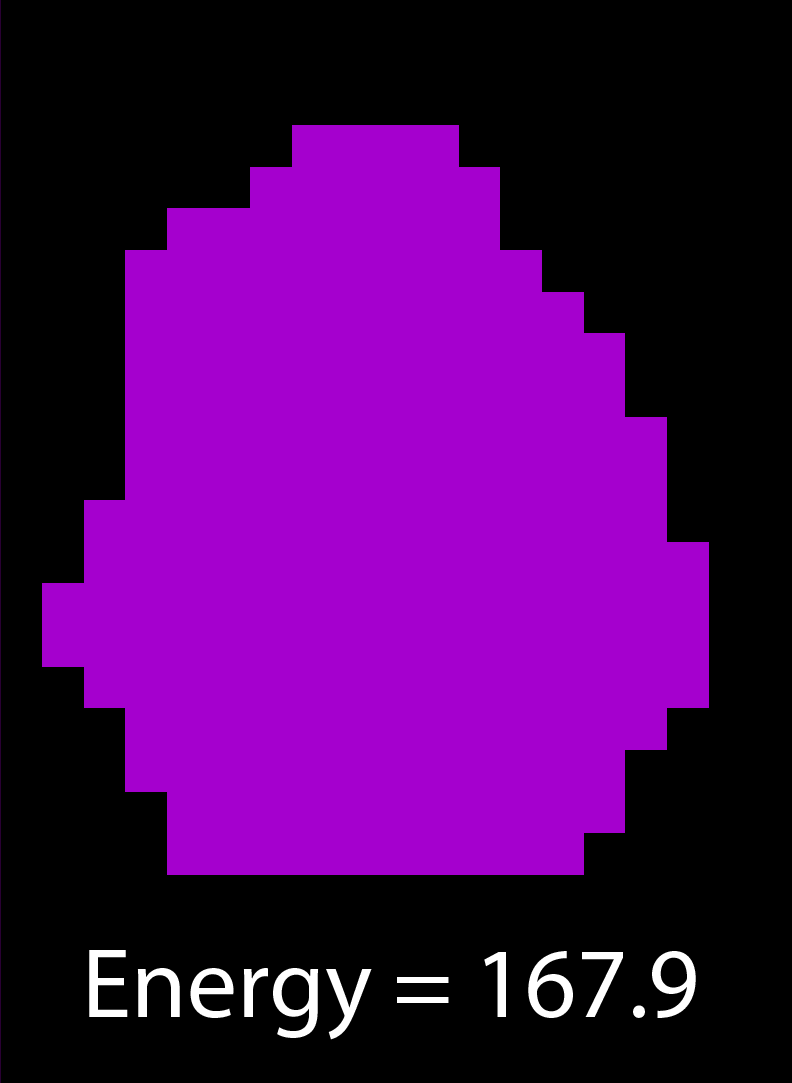} \\
	(a) & (b) & (c) & (d) & (e) 
\vspace{-5pt}
\end{tabular}
\end{center}
   \caption{ Comparison of our method to the state-of-the-art method of Gorelick et al.\ \cite{BoykovECCV2014} on a synthetic and a biological image. 
   Top row: (a)~Synthetic image. (b) Optimal solution of a Potts model. (c,d) Method~\cite{BoykovECCV2014}, initialized by (b).  (c) First iteration, and (d) sixth iteration (converged), with energy. (e)~Our method obtains the globally optimal convex solution. 
   Bottom: (a)~Biological image (cf.\ Fig.\ \ref{fig:2LabelPotts}a). (b)~Optimal solution of a Potts model. (c)~Solution of method~\cite{BoykovECCV2014}, initialised by (b), and (d) initialized by a square manually placed around the sought structure (top right small image). Both solutions of~\cite{BoykovECCV2014} are sub-optimal. (e) Our method yields a globally optimal convex solution.  
   }
\label{fig:BoykovComparison}
\label{fig:boykov-bio}
\end{figure*}

\textbf{Potts Model. }
%
Figure~\ref{fig:2LabelPotts} shows examples to which we apply two-label Potts models enriched by convexity constraints on the foreground label by means of the Convex-MCN problem~\ref{def:multiway-cut-convex}.
Figure~\ref{fig:2LabelPotts}(a) 
shows a biological image of ellipsoidal \emph{cell nuclei} in the tissue of a nematode worm.
The image shows four whole nuclei, and a fraction of a fifth one at the bottom. 
Convexity constraints allow for a perfect segmentation of the nuclei. In contrast, a Potts model without convexity constraints fails to split apart the nuclei, and yields holes within components. 
Figure~\ref{fig:2LabelPotts}(d) 
shows a biological image of polygonal cells in a fly wing. Again, convexity constraints allow for a perfect segmentation of the nuclei, while a respective Potts model without convexity constraints is not able to correctly split apart the cells.
Figure~\ref{fig:3LabelPotts}(a) 
shows a photograph to which we apply a three-label Potts model with convexity constraints on one foreground label. 
Our method is able to accurately segment two convex foreground objects on top of a second, non-convex foreground label. 
%
%

Results in Figures~\ref{fig:2LabelPotts} and~\ref{fig:3LabelPotts}(c)
were achieved with ``simple'' convexity constraints as captured by~\eqref{eq:convexity-constraints-multiway-as-implication}. 
An example for the more general constraints~\eqref{eq:convexity-constraints-relative-multiway-as-implication} is given in Figure~\ref{fig:3LabelPotts}(d-f). 
Here, we enforce components labeled ``apple'' to be convex as such, while we enforce components labeled ``orange'' to be convex only w.r.t.\ the background. 
In other words, we allow concavities in orange-labeled components as long as they are filled exclusively by apple-labeled nodes. 
Consequently such concavities appear as desired. However, incorrect spurious apple components also appear within orange components. 
This result is interesting despite the fact that our model does not achieve a perfect segmentation, because it shows potential implications of the generalized constraints. 
%


\textbf{Correlation Clustering. }
Figure~\ref{fig:CorrelationClustering} shows two exemplary results for correlation clustering enriched by convexity constraints by means of the Convex-MC problem~\ref{def:multicut-convex}. 
In both examples, densely packed objects, namely bricks in a stone wall and cells in a fly wing, are nicely separated due to convexity constraints. At the same time, the space between these objects is tesselated into convex components. 
Although these results might not be of direct use as segmentations of the respective objects, they may well serve as convex supervoxelizations to be used as input for further processing. 
%



\label{results:subsec:boykov} 
\paragraph{Comparison to State-Of-The-Art on Two Exemplary Images. }
We compare our method to the state-of-the-art for segmentation with convexity constraints~\cite{BoykovECCV2014}. 
The method of Gorelick et al.\ \cite{BoykovECCV2014} is able to handle only one convex structure of one foreground label, and we chose exemplary images accordingly (Figure~\ref{fig:BoykovComparison}). We use the code provided by the authors. 
First we study a synthetic image (Figure~\ref{fig:BoykovComparison} top). The method of~\cite{BoykovECCV2014} initializes via the Graph Cuts solution, yielding a hole in the foreground object. In the process of resolving this high energy configuration, the method breaks the outer boundary of the object and settles to a sub-optimal solution. 
In contrast, our method is able to obtain the globally optimal solution. 
On a second, biological image (Figure~\ref{fig:BoykovComparison} bottom), we ran~\cite{BoykovECCV2014} with two different initializations, namely the standard Graph Cut solution, as well as a box that we manually placed around the sought structure. Both initializations result in sub-optimal solutions, whereas our method is again able to obtain a globally optimal solution.  
However, we note that the method of~\cite{BoykovECCV2014} is considerably faster than ours. 

%
%
%

\section{Conclusion and Future Work}

We proposed a new approach that introduces convexity constraints into two multicut problems that are equivalent to Correlation Clustering and the Potts Model, respectively. Our approach handles convexity constraints for many connected components of multiple different classes, and additionally for pre-specified convexity relationships between objects of different classes. All concepts described in this paper extend in a straightforward way to 3D. In future work we will explore strategies for improving the run-time, e.g.\ via warm-start of subsequent ILPs, for the application to larger data.

\section*{Appendix}
  \paragraph{Proof of Theorem~\ref{theo:1}. }
  \label{app:1}
  First we show that every multicut satisfies $| C \cap E_\Pi | \neq 1$ for all cycles. 
Given a decomposition and related multicut $E_\Pi$, assume there exists a cycle $C$ with $C \cap E_\Pi = \{uv\}$. Then $uv$ straddles distinct components because $uv \in E_\Pi$. On the other hand, the path $C \setminus \{uv\}$ connects $u$ and $v$ with edges that are not in $E_\Pi$, and hence $u$ and $v$ belong to the same component. By contradiction, this proves that $| C \cap E_\Pi | \neq 1$ for all cycles. 
Second we show that every $Y$ that satisfies $| C \cap Y | \neq 1$ for all cycles is a multicut of $G$. 
Given such a $Y$, we construct a decomposition $\Pi$ and related multicut $E_\Pi$ of $G$ as follows: 
$uv \not\in E_\Pi :\Leftrightarrow$ there exists a path $P$ between $u$ and $v$ such that $P \cap Y = \emptyset$. 
Now we show that $Y=E_\Pi$. 
If $uv\not\in Y$, then $uv\not\in E_\Pi$, because there is a path between $u$ and $v$ such that $P \cap Y = \emptyset$, namely $P=\{uv\}$.
If $uv\not\in E_\Pi$, assume $uv\in Y$. Then, because $uv\not\in E_\Pi$, there exists a path between $u$ and $v$ such that $P \cap Y = \emptyset$.
Hence the cycle $C:=P\cup\{uv\}$ satisfies $|C \cap Y | = 1$. By contradicting $uv\in Y$, this proves that if $uv\not\in E_\Pi$, then $uv\not\in Y$. 
\paragraph{Proof of Lemma~\ref{lemma:1}. }
\label{app:2}
Given a path $P$. 
In case $\sum_{e\in P} y_{e} = 0$ it follows from $y_{e}\geq 0$ that  
\eqref{eq:mc-convexity} 
$\Leftrightarrow \sum_{e\in S(P)} y_{e} \leq 0 \Leftrightarrow \sum_{e\in S(P)} y_{e} = 0 \Leftrightarrow$ \eqref{eq:convexity-constraints-as-implication}.
Case $\sum_{e\in P} y_{e} \neq 0$ 
entails $\sum_{e\in P} y_{e} \geq 1$ 
because
$y_{e}\in\{0,1\}$.
Hence
$|S(P)| \cdot \sum_{e\in P} y_{e}\geq |S(P)|$. 
And, 
$|S(P)| \geq \sum_{e \in S(P)} y_{e}$ 
because 
$y_{e}\leq 1$. 
So, 
$true \Leftrightarrow$ 
\eqref{eq:mc-convexity} 
$\Leftrightarrow$ 
\eqref{eq:convexity-constraints-as-implication}.
%
%

\clearpage
{\small
\bibliographystyle{ieee}
\bibliography{convexMulticut2015}
}

\end{document}